\documentclass[final,3p,times,twocolumn]{elsarticle}
\usepackage{amsmath}

\usepackage{multirow}
\usepackage{amssymb}
\usepackage{booktabs}
\usepackage{tabularx}

\journal{Neurocomputing}

\begin{document}

\begin{frontmatter}


\author{Ruiqi Lu\fnref{label1}}
\author{Huimin Ma\corref{cor1}\fnref{label2}}

\cortext[cor1]{Corresponding author}
\ead{mhmpub@ustb.edu.cn}

\title{Semantic Head Enhanced Pedestrian Detection in a Crowd}

\address[label1]{Department of Electronic Engineering, Tsinghua University}
\address[label2]{School of Computer \& Communication Engineering, Institute of Artificial Intelligence,\\ University of Science \& Technology Beijing}

\begin{abstract}
Pedestrian detection in the crowd is a challenging task because of intra-class occlusion. More prior information is needed for the detector to be robust against it. Human head area is naturally a strong cue because of its stable appearance, visibility and relative location to body. Inspired by it, we adopt an extra branch to conduct semantic head detection in parallel with traditional body branch. 
Instead of manually labeling the head regions, we use weak annotations inferred directly from body boxes, which is named as `semantic head'. In this way, the head detection is formulated into using a special part of labeled box to detect the corresponding part of human body, which surprisingly improves the performance and robustness to occlusion.
Moreover, the head-body alignment structure is explicitly explored by introducing Alignment Loss, which functions in a self-supervised manner.
Based on these, we propose the head-body alignment net (HBAN) in this work, which aims to enhance pedestrian detection by fully utilizing the human head prior. Comprehensive evaluations are conducted to demonstrate the effectiveness of HBAN on CityPersons dataset.
\\
\\
\textit{Keywords}: Pedestrian Detection; Occlusion; Semantic Head; Head-Body Alignment
\end{abstract}





\end{frontmatter}


\section{Introduction}
\label{sec1}
Pedestrian detection is a key component for many computer vision tasks such as autonomous driving, person re-identification and surveillance. Recent deep CNN based methods \cite{repulsion,illuminating,improve-handling,asymtotic-fitting,What-help} have led to great advances in pedestrian detection. However, it remains a challenging problem as pedestrians often occlude each other. Detectors should be able to detect pedestrians with heavy occlusion and discriminate different objects in a crowd.

Human body parts have been proved to be useful in state-of-the-art detection models \cite{multi-label,strong-parts,attention,ORCNN,PCN} to deal with occlusion. However, as there exists no part annotations in common pedestrian detection datasets \cite{kitti,citypersons,caltech}, these methods either use extra information (eg. key parts or human parsing) or manually divide the complete body box into grids and define parts as different combinations of several grids. As a result, they are not able to construct part detectors to generate part detections and the restriction between human parts is only implicitly embedded in the classification stage.  

\begin{figure}
	\centering
	\includegraphics[width=1.00\linewidth]{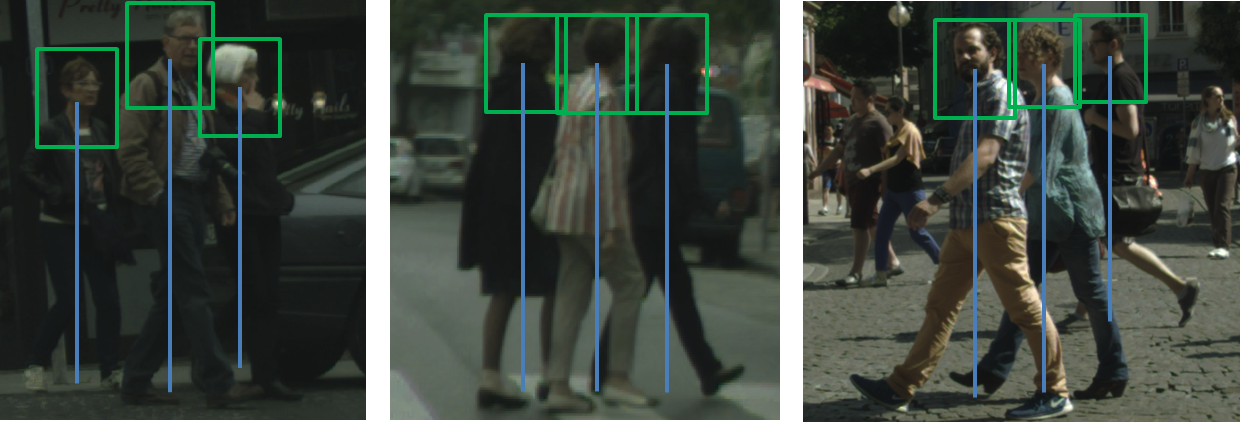}        
	\caption{Illustration of head regions of different pedestrians. Head is an important cue for detection in the crowd because it's hard to be occluded, stable in appearance and located stably at the top center of a standing pedestrian.}
	\label{head demonstration}
\end{figure}

Different from these methods, the model proposed in this paper directly constructs part detectors by training from weak annotations and the model is deeply coupled with FRCNN \cite{faster} framework. The difficulty of training part detectors is that under crowded scenes, semantic parts like legs and hands are often occluded, and it's difficult to discriminate which pairs of hands or legs belong to which objects. In addition, the relative location between hands or legs and body is too unstable to learn a robust part detector. Compared with other parts of human, the head area is located stably near the top center of a standing pedestrian with relatively stable appearance, as shown in Figure \ref{head demonstration}. Furthermore, head area is hard to be occluded by other heads or objects normally, due to its relative high location. Hence, constructing a head detector is a feasible way of employing part information in pedestrian detection.

Inspired by it, a head-body alignment net (HBAN) based on the FRCNN framework is proposed in this paper to address the occlusion problem. Specifically, the semantic head and body boxes are produced in parallel and are required to align with each other to produce accurate detections through a new loss function named alignment loss (AlignLoss).

In order to utilize head information in a CNN-based detector, massive head annotations are needed for training, which are not provided in common pedestrian datasets. However, we find it unnecessary to do so because the head-body structure is usually stable for a standing pedestrian, as shown in Figure \ref{head demonstration}. In addition, the annotation protocol adopted for some datasets like CityPersons \cite{citypersons} is strict, where the full body is annotated by drawing a line from the top of the head to the middle of two feet, and the bounding box is generated using a fixed aspect ratio(0.41). Therefore, the head region is usually located at the top center of a body box and annotations can be automatically inferred from body boxes, see Figure \ref{head definition}. Because the generated annotations contain not only the human head, but also shoulder area, we refer the region as `semantic head' (s-head for short) to avoid misunderstanding.  

We make the following three contributions in this work:

\noindent \textbf{1.} We propose to detect semantic head boxes by designing s-head RPN and RCNN in parallel with normal body branch. The annotations for training the s-head detector are inferred directly from body boxes by using the location prior. This way, the head detection is formulated into using a special part of labeled box to detect the corresponding part of human body.

\noindent \textbf{2.} We propose AlignLoss, which explicitly requires the body box to locate compactly around it's s-head region to produce accurate detections.

\noindent \textbf{3.} We conduct comprehensive experiments with different training settings, which demonstrate the effectiveness of semantic head in improving detection performance, especially for occluded pedestrians.

\section{Related Works}
Our work is focused on pedestrian detection on road scene and deals with occlusion by incorporating s-head detection. Therefore, we review recent work on CNN based pedestrian detectors, occlusion handling for pedestrians and human head detection methods, respectively.

\subsection{Pedestrian Detection}
Among different detection tasks, pedestrian detection has its significance for many real-world applications like autonomous driving. Multiple benchmarks \cite{caltech,citypersons,kitti} have been proposed with challenges including small scale, pose variations and occlusion. Common deep models \cite{R-FCN,FPN,focal-loss} designed for general object detection are not naturally suitable for pedestrian detection and need proper modifications. RPN+BF \cite{RPN+BF} states that the second stage regressor degrades the results because of insufficient resolution. By replacing the second stage classifier with boosted forests, it is able to boost the performance. Adapted FRCNN \cite{citypersons} revisits CNN design and points out key adaptations including finer feature stride and ignore region handling to enable plain FRCNN to obtain state-of-the-art results. SAF \cite{scale-aware} builds two subnetworks for pedestrians of large or small scales.  Segmentation information \cite{small,fusedDNN,illuminating} is adopted to enable convolution layers to learn more robust and discriminative features. ALFNet \cite{asymtotic-fitting} proposes asymptotic localization fitting to evolve the default anchor boxes of SSD by steps into improving results. CSP \cite{CSP} adapts recently proposed anchor-free models into pedestrian detection by directly predicting the center and scale of pedestrians.

\subsection{Occlusion Handling}
One of the challenges for pedestrian detection is intra-class occlusion. Many works handle occlusion by introducing specially designed loss, exploiting the unique body structure of human or refining post-processing NMS operation.

Repulsion loss \cite{repulsion} is designed to directly penalize the predicted box for shifting to the other ground-truth objects or other predicted boxes associated with different ground truths. Soft NMS \cite{soft-NMS} decays the detection scores of boxes depending on the overlap and eliminates no boxes in the process to achieve high recall. Adaptive NMS \cite{adaptive-NMS} applies a dynamic suppression threshold according to the target density. 
NMS network \cite{NMS} relies on an additional post-processing network to learn NMS and achieve the function of duplication removal. Bi-Box \cite{bi-box} also adopts a two branch detection framework. It tries to detect the visible box of a pedestrian in the other branch while we try to detect the head region. 

Apart from these methods, human part information is most widely used in occlusion handling.
DeepParts \cite{strong-parts} introduces DPM \cite{DPM} model into deep network for pedestrian detection.
Occlusion-aware detection score is proposed in \cite{improve-handling} where the classifier of a single stage detector not only gives body scores but also produces part grid confidence map for an anchor box. 
Similarly, PCN \cite{PCN} also divides the proposals into fixed grids and produces confidence maps, but it uses LSTM \cite{LSTM} to process different permutations of part scores as sequences. 
OR-CNN \cite{ORCNN} replaces the RoI pooling layer with a Part Occlusion-aware RoI Pooling unit to integrate the prior structure of human body into the network.
Part attention maps produced by key-point detection model are adopted to re-weight convolution features in \cite{attention}.
Because human part annotations are not provided in common datasets, most of part-based methods avoid constructing part detectors by manually defining part prototypes or different combinations of grids. 
Our method is close to part-based methods stated above. But we only consider semantic head as additional part information and train a part detector for semantic head directly and couple it into both RPN and RCNN stages. Moreover, the restriction between semantic head and body is explicitly utilized by introducing a self-supervised loss, named Alignment Loss.

\begin{figure*}
	\centering
	\includegraphics[width=1.00\linewidth]{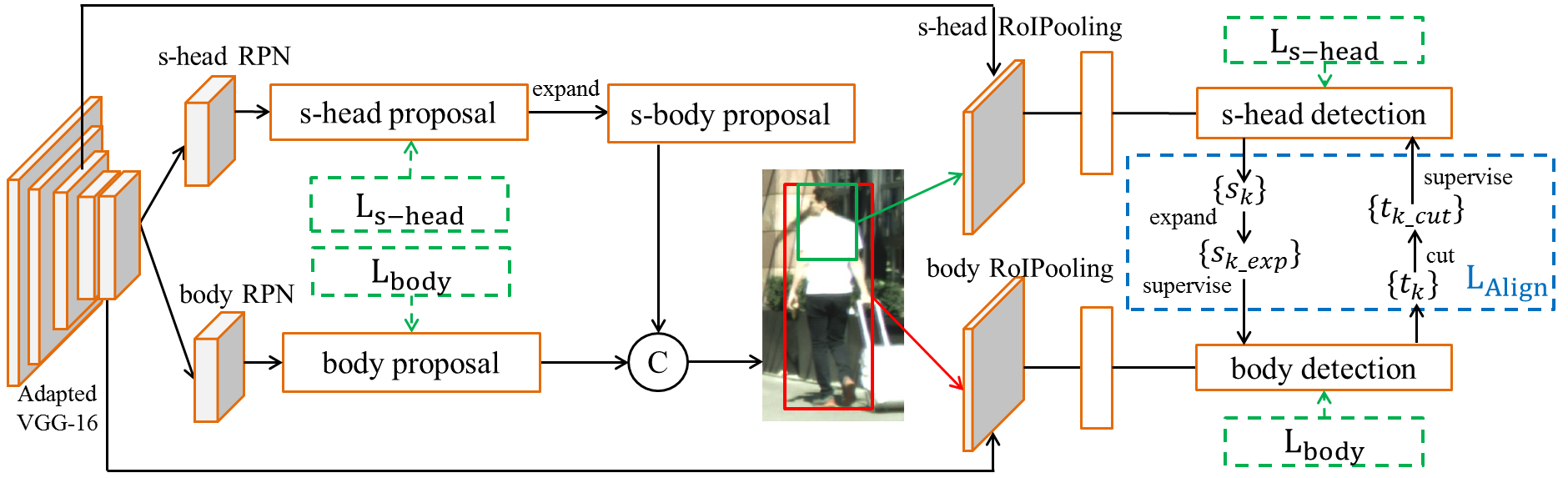}        
	\caption{Overview of HBAN framework. Both RPN and RCNN stages adopt one branch for s-head and another for body, which are trained with separate loss (dashed green). RCNN stage is also trained with AlignLoss (dashed blue) to ensure that body and s-head boxes align with each other, which is also demonstrated. The s-head proposals are expanded and combined with body proposals for the next stage. NMS will be conducted on body detections to produce final results. }
	\label{overview}
\end{figure*}

\begin{figure}
	\centering
	\includegraphics[width=1.0\linewidth,height=0.35\linewidth]{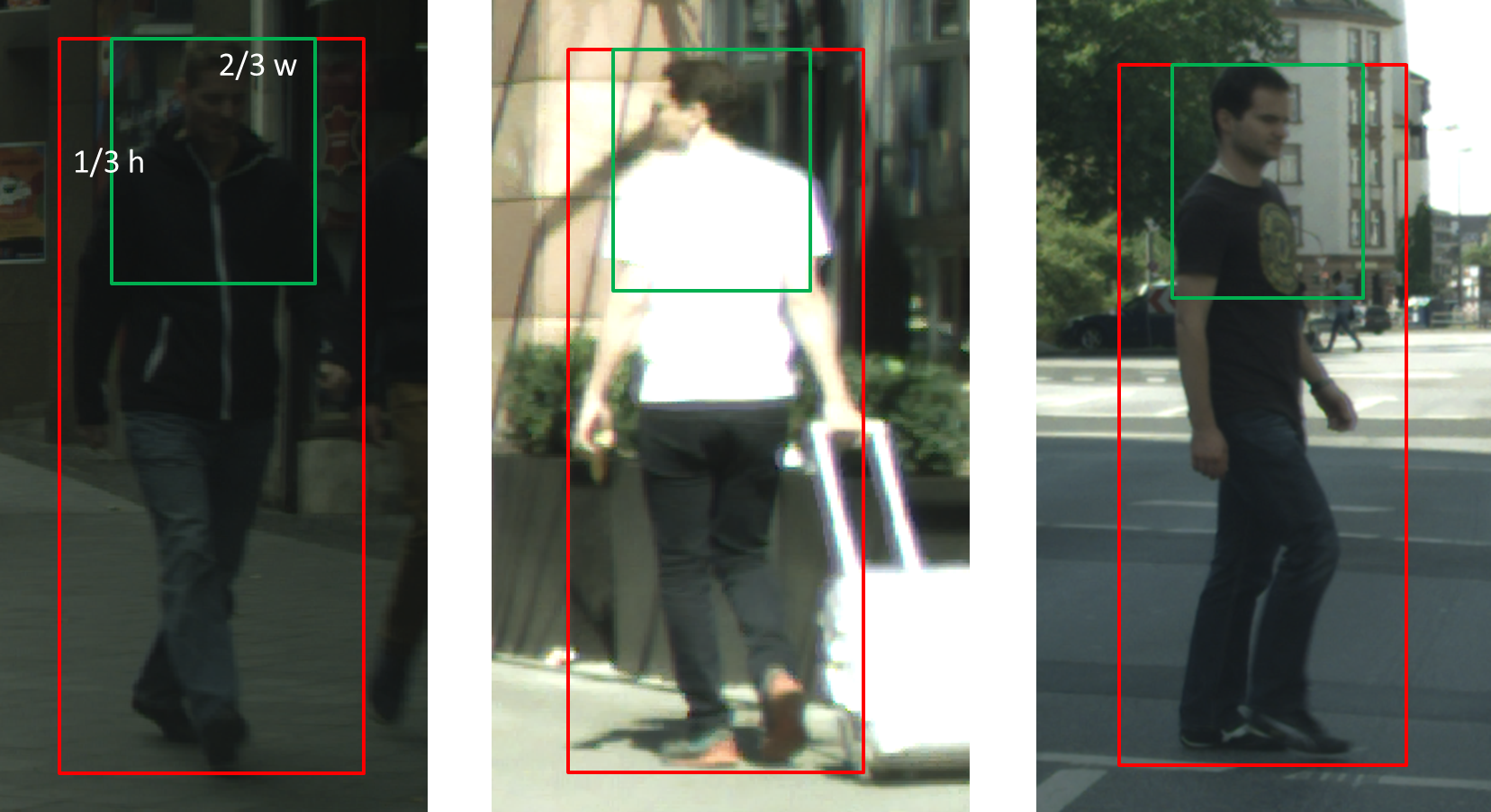}        
	\caption{The definition of semantic head region. The upper $\frac{1}{3}$ and middle $\frac{2}{3}$ of the body box is regarded as its s-head region, which contains the head and shoulder area.}
	\label{head definition}
\end{figure}
\subsection{Head Detection}

Head detection is a key component for counting the number of people under crowded scenes. Compared with other parts of body, head is stable in appearance and is hard to be occluded. Schmid \cite{head-1} uses DPM template method to get upper body proposal first and then refine it to the head location. CNN classifier is introduced in \cite{head-2} combining head and context information. FRCNN with head-shoulder context is adopted in \cite{lechao} to bring more effective context information. All these methods use real head annotations for training, which are not provided in pedestrian detection datasets. Compared with them, we automatically generate semantic head annotations from full body annotations, using the prior that head region usually aligns with the top center of a standing pedestrian. Thus, we are able to train head detector with no extra labeling information. It also transforms the task into detecting fixed regions of body using fixed annotations, which is shown to be helpful for full body detection.

\section{Head-Body Alignment Net} 

\subsection{Adapted FRCNN}
The proposed architecture is based on the Adapted FRCNN framework \cite{attention} for pedestrian detection. It refines on the plain FRCNN by proposing the following 3 modifications, which are followed in this work.

\noindent1. Quantized RPN scales. It split the full scale range of samples in 10 quantile bins (equal amount of samples per bin) and use the resulting 11 endpoints as RPN scales to generate proposals. The ratio of all anchors are set to                      $2.44$ $(\frac{1}{0.41})$.

\noindent2. Finer feature stride. It removes the fourth max-pooling layer from VGG16 and reduces the stride to 8 pixels, helping the detector to handle small objects.

\noindent3. Ignore region handling. The ignore region annotations include areas where the annotator cannot tell if a person is present or absent, and person groups where individuals cannot be told apart. During training the RPN and RCNN, proposals and RoIs avoid sampling the ignore regions.

\subsection{Framework}
 The framework consists of a region proposal network (RPN) to generate candidate bounding boxes and corresponding scores, and a regression network (RCNN) to refine the boxes and scores. In both stages, s-head boxes are produced in parallel with the full body region. To use s-head as a structural restriction for body, AlignLoss is proposed to enforce the s-head regions to align with the top center of body in a self-supervised manner. Figure \ref{overview}. shows the overview of proposed framework. Because the s-head region is much smaller than body, for s-head RoIPooling layer, the feature maps output by conv3 block are pooled instead of conv5 block (the backbone is VGG-16).

\subsection{Semantic Head}
Following \cite{lechao}, the definition of s-head box contains the head and shoulder area, shown in Figure \ref{head definition}. Quantitatively, assume $B=(x_1,y_1,x_2,y_2)$ represents the coordinates of upper left and lower right corners of body. Then its corresponding s-head box is $H=(x_1+\frac{1}{6}w, y1, x_2-\frac{1}{6}w, y1+\frac{1}{3}h)$, where $w$ and $h$ represent the width and height of body box. This means that the upper $\frac{1}{3}$ and middle $\frac{2}{3}$ of the body box is regarded as its s-head region. The scale of s-head anchor is therefore set to be $2/9$ of body anchor to match its size. By reversing the above calculation, we could also expand a s-head box to its body region. 

In the first stage, a two-branch RPN architecture is adopted, where one branch is responsible for the full body region, and the other branch is responsible for s-head. These two branches share convolution layers and produce separate box and score results. By expanding the s-head proposals to its body region, proposals generated from both branches are re-ordered by their scores and combined together after NMS for the next stage. In the second stage, a similar parallel RCNN architecture is adopted. Separate RoI pooling layers for s-head and body regions give separate s-head and body detections of the same pedestrian. 

\begin{figure}
	\centering
	\includegraphics[width=1\linewidth,height=0.45\linewidth]{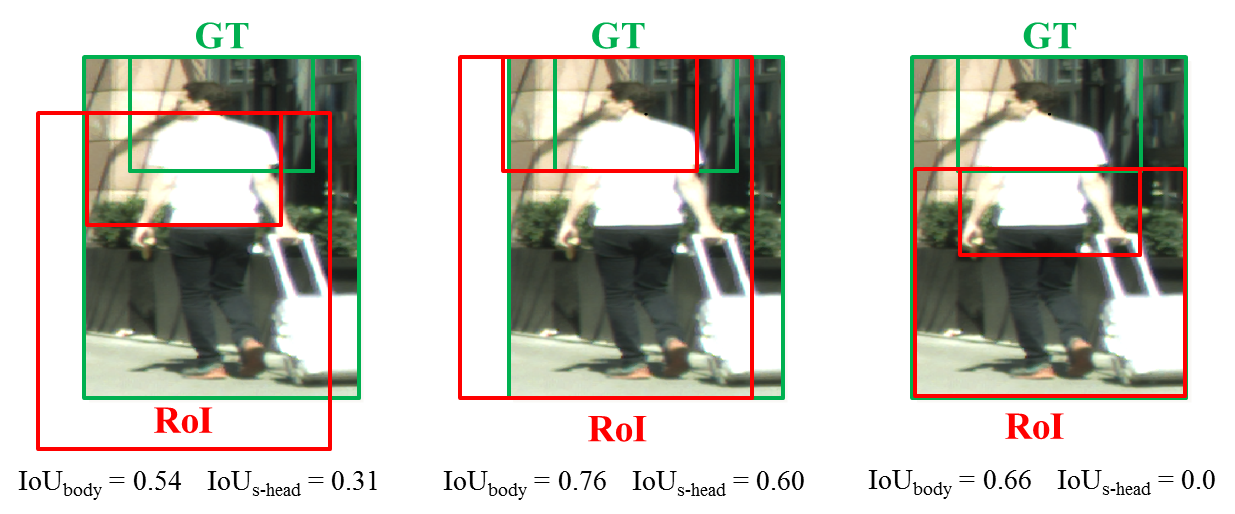}        
	\caption{Demonstration of selecting training samples. The labels of body and s-head may not be identical.}
	\label{fig:positive-demon}
\end{figure}

While the body branch is supervised by labeled full body annotations $B$, the s-head branch is supervised by inferred regions $H$, whose location is fixed relative to $B$. As no extra labeled information is introduced in the framework, one may wonder why the extra s-head branch would help increase the performance. During training, the positive body and s-head samples are selected independently. Demonstrated in Figure \ref{fig:positive-demon}, when the full body of a RoI has IoU larger than a predefined threshold (0.5 for example) with ground truth full body, the s-head of it may not have the same IoU with ground truth s-head. Therefore, the label of body and s-head may be conflict for the same RoI. Such difference in discriminating positive and negative samples will guide the network to focus more on the head region to determine whether the RoI is a pedestrian and the multi-task training target has the potential of increasing the detecting performance, which is verified in experiments.

\subsection{Alignment Loss}
For a standing pedestrian, the s-head region is well aligned with the top center of its bounding box, while the top center of some false positives contain other body parts or background. Therefore, by enforcing s-head boxes to align with the top center of the body boxes, we can get more accurate detections. In order to explicitly apply the relationship, AlignLoss is designed for the RCNN module, which enforces body box to locate compactly around its s-head region. For RCNN stage, the total loss is defined as:
\begin{equation}
L_{RCNN} = L_{body} + L_{s-head} +　L_{Align}
\end{equation}
where $L_{body}$ and $L_{s-head}$ terms are the same as in conventional FRCNN, which consist of classification loss and regression loss requiring each RoI (body or s-head) to approach the target ground truth. They are defined as:
\begin{equation}
\begin{split}
L_{body}(&\{p_i\},\{t_i\},\{p_i^*\},\{t_i^*\})=\\&L_{cls}(\{p_i\}, \{p_i^*\}) + \alpha\cdot L_{reg}(\{p_i^*\},\{t_i\},\{t_i^*\})
\end{split}
\end{equation}
\begin{equation}
\begin{split}
L_{head}(&\{q_i\},\{s_i\}, \{q_i^*\},\{s_i^*\})=\\&L_{cls}(\{q_i\}, \{q_i^*\}) + \alpha\cdot L_{reg}(\{q_i^*\},\{s_i\},\{s_i^*\}) 
\end{split}
\end{equation}
where $i$ is the index of RoIs in a batch, $p_i$ and $t_i$ are the classified confidence and regressed coordinates of the $i$-th body RoI, and $p_i^*$ and $t_i^*$ are the corresponding ground truth body label and coordinates. Similarly, $q_i$, $s_i$, $q_i^*$ and $s_i^*$ are for s-head RoIs.  $L_{cls}$ is the SoftmaxLoss and $L_{reg}$ is the SmoothL1Loss.

The $L_{Align}$ term in (1) is the AlignLoss which enforces the s-head box to locate at the top center of corresponding body box of the same pedestrian. Therefore, it considers both body and s-head results. Specifically, we denote $t_{cut}$ as the virtual s-head region cut from corresponding body box $t$, and ${s_{exp}}$ as the virtual body region expanded from s-head box $s$. The cut and expand procedure follows the definition of $B$ and $H$ explained in Section 3.2. 

As the s-head result needs to align with body in space, it's reasonable to require the virtual s-head region $t_{cut}$ cut from $t$ to be close to the s-head result $s$ associated with the same pedestrian. Symmetrical requirement for $s_{exp}$ and $t$ is also true. Therefore, the $L_{Align}$ is defined as:
\begin{equation}
L_{Align} = \frac{1}{|P|}\sum_{s\in P}[\Delta(t,\,s_{exp}) + \Delta(s,\,t_{cut})]
\end{equation}                                                                                                             
where $P$ is a batch of RoIs whose s-head labels are positive and $\Delta(\cdot)$ is the SmoothL1loss. 

From the definition above, the virtual s-head region cut from body RoI $t_{cut}$ is compared with $s$, which means that $t_{cut}$ is regarded as `ground truth' for s-head branch. Symmetrically $s_{exp}$ is also regarded as `ground truth' for body branch. As the real ground truth is not utilized here, AlignLoss is a kind of self-supervised loss, where two branches restrict each other, making semantic head and body boxes to locate compactly with each other. 

AlignLoss is not considered in RPN stage. The main reason is that in RCNN stage the s-head and body results of the same pedestrian are produced in pairs, while in RPN stage it is unclear which anchors from body RPN should match those in s-head RPN. A possible way is taking the mean coordinates of all positive anchors corresponding to the same pedestrian. But such operation is time consuming and the accuracy of mean box relies on the IoU threshold adopted in RPN. Therefore, we only conduct AlignLoss in RCNN stage.

\section{Experiments} 
\subsection{Dataset and Evaluation Metric}
\noindent \textbf{Dataset.} We evaluate the proposed HBAN on CityPersons \cite{citypersons}, which is a challenging dataset for occluded pedestrian detection and exhibits large diversity. It comes from the semantic segmentation dataset 
CityScapes \cite{cityscape} and consists of 5,000 images captured in 18 cities in Germany. Both full and visible box annotations for 2975 train and 500 validation images are provided. The test set is only used for online evaluation. The total number of annotated pedestrians is 19,654 in the training set, among which about 56\% pedestrians are with `reasonable' occlusion and 30\% ones are heavily occluded.  

\noindent \textbf{Evaluation Metric.} The log miss rate averaged over the false positive per image (FPPI) range of [10$^{-2}$, 10$^0$] (MR$^{-2}$) is used (lower is better). CityPersons dataset is split into different subsets according to the occlusion ratio and height of pedestrians. Following \cite{citypersons}, we report results on Reasonable (\textbf{R}) and Heavy (\textbf{HO}) subsets. The occlusion ratio in \textbf{R} is smaller than 35\% and the occlusion ratio in \textbf{HO} is between 35\% and 80\%. Only pedestrians with height over 50 pixels are considered in the subsets.

\subsection{Implementation Settings}
\noindent \textbf{Training sets.}   The Reasonable subset (\textbf{R}) is commonly used in training \cite{ORCNN, repulsion, asymtotic-fitting}. However, as we focus on detecting pedestrians with various occlusion levels, different subsets are used in the experiments to train HBAN in order to provide a thorough evaluation. Following \cite{bi-box, mask-guided}, we also use another training set which contains pedestrians with occlusion ratio less than 70\%, denoted as \textbf{R+}.

\noindent \textbf{Training Details.} We train for 20k iterations first with the base learning rate of 0.001 and another 10k iterations with learning rate decreased by 10. When training on \textbf{R+} set, we also adopt visible ratio to sample positive RoIs, as introduced in Bi-Box \cite{bi-box}. The SGD solver and a mini-batch of 2 images is adopted to optimize the parameters on a single Titan X GPU. Online hard example mining (OHEM) is adopted for better convergence. 

\subsection{Is Head the Optimal Part?}
Human head prior is considered to be stable and robust in this paper and the detector for semantic head is trained through end-to-end back propagation. It enables us to produce s-head box rather than manually defining it as certain combination of grids. Yet the robustness of head is so far an empirical saying. Therefore, we validate it by experiments.

\begin{table}
	\begin{center}
		\renewcommand\arraystretch{1.1}
		\setlength{\tabcolsep}{5mm}{ 
			\begin{tabular}{@{}lccc@{}}
				\hline
				Part & \textbf{R} & \textbf{HO} & \textbf{R \& HO}\\
				\hline
				S-Head & \textbf{25.59} & \textbf{56.52} & \textbf{50.76}\\
				Middle & 33.88 & 76.63 & 61.42\\
				Lower & 43.35& 82.80 & 67.47\\
				Left & 33.81  & 72.70 & 60.42\\
				Right & 36.14 & 73.59 & 61.67\\
				\hline
			\end{tabular}
		}
	\end{center}
	\vspace{-2mm}
	\caption{Performance of different single part detectors evaluated by MR$^{-2}$ score (lower is better) on CityPersons validation set.}
	\label{table:single part detector}
\end{table}

Notice that either branch (s-head or body) in Figure \ref{overview} can fulfill pedestrian detection task. For the body branch, the proposal or regression output is naturally the desired result. For the s-head branch, the full body box can be obtained by expansion. Therefore, we could disable the body branch and only use the s-head to detect pedestrians. Moreover, to validate the robustness of human head, we could switch s-head to other body parts and compare their performances. 

\begin{figure}
	\centering
	\includegraphics[width=1.0\linewidth,height=0.35\linewidth]{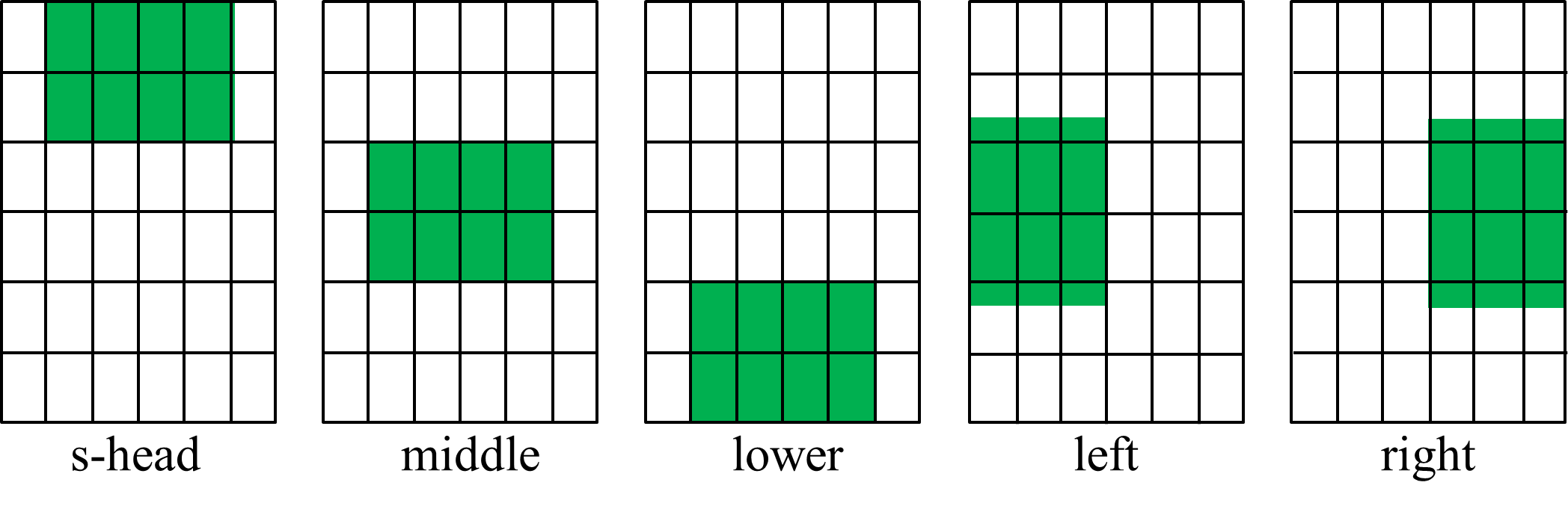}   
	\vspace{-2mm}     
	\caption{The part pool. Green regions denote the parts. Every part contains equal area for fair comparison.}
	\label{fig:single part definition}
\end{figure}

The experiments are conducted on CityPersons using \textbf{R} set for training. The part pool contains s-head, middle, lower, left and right of human body, which are commonly used in part-based methods. Their definitions can be seen in Figure \ref{fig:single part definition} and the area of each part is equal for fair comparison. We separately train a detector for each part by cutting the corresponding region as weak supervision and train part RPN and RCNN in an end-to-end way. The anchor settings are adjusted accordingly. Then during test phase, NMS is conducted directly on part detections before expanding to the full body as the final output. Performances are demonstrated in Table \ref{table:single part detector}. Using s-head as prior surpasses all other parts with a large margin (8.2\% at least), which proves the robustness of head prior. It's possible to integrate all parts into one framework, i.e. constructing a branch for each part. However, the number of parameters will grow large and it's difficult to determine the loss weight of each branch. More importantly, other parts are more likely to hurt the performance as they carry much more noise and the head prior is specially chosen in this paper for its alignment with body.

\subsection{Ablation Studies}
The ablation studies are conducted using two training sets. First is the Reasonable set (\textbf{R}), which contains pedestrians with height $\ge 50$ and occlusion $\le 0.35$. It is commonly used by most methods. But as we focus on pedestrians with various occlusion levels, training on \textbf{R} is insufficient to thoroughly evaluate our model. Therefore, following Bi-Box \cite{bi-box}, we use another training set, which contains pedestrians with height $\ge 50$ and occlusion $\le 0.7$, denoted as \textbf{R+}.
The results are demonstrated in Table \ref{ablation study:0.65} and Table \ref{ablation study:0.3}, respectively. The overall performance on Reasonable and Heavy subsets can reflect a model's robustness against occlusion.

\begin{table}
	\begin{center}
		\renewcommand\arraystretch{1.1}
		\setlength{\tabcolsep}{1.5mm}{ 
			\begin{tabular}{@{}lccccc@{}}
				\hline
				Method & 2RPN & 2RCNN & AlignLoss & \textbf{R} & \textbf{HO} \\
				\hline
				Baseline & & & & 14.28 & 53.24 \\
				\hline
				HBAN & \checkmark & & & 13.68 & 49.61  \\
				&  & \checkmark &   & 13.81 & 50.07\\
				&  & \checkmark & \checkmark  & 12.90 & 48.45\\
				& \checkmark & \checkmark & \checkmark  & \textbf{12.51}  & \textbf{48.07}\\
				\hline
			\end{tabular}
		}
	\end{center}
	\vspace{-2mm}
	\caption{Ablation studies of HBAN on CityPersons validation set. Models are trained using \textbf{R}. }
	\label{ablation study:0.65}
\end{table}

\begin{table}
	\begin{center}
		\renewcommand\arraystretch{1.1}
		\setlength{\tabcolsep}{1.5mm}{ 
			\begin{tabular}{@{}lccccc@{}}
				\hline
				Method & 2RPN & 2RCNN & AlignLoss & \textbf{R} & \textbf{HO}  \\
				\hline
				Baseline & & & & 14.74 & 46.69 \\
				\hline
				HBAN & \checkmark & & & 13.95 & 44.25  \\
				&  & \checkmark &   & 13.87 & 45.30 \\
				&  & \checkmark & \checkmark  & 13.76 & 43.32\\
				& \checkmark & \checkmark & \checkmark  & \textbf{12.99}  & \textbf{42.39}\\
				\hline
			\end{tabular}
		}
	\end{center}
	\vspace{-2mm}
	\caption{Ablation studies of HBAN on CityPersons validation set. Models are trained using \textbf{R+}. }
	\label{ablation study:0.3}
\end{table}

\begin{figure}
	\centering
	\includegraphics[width=0.75\linewidth,height=0.5\linewidth]{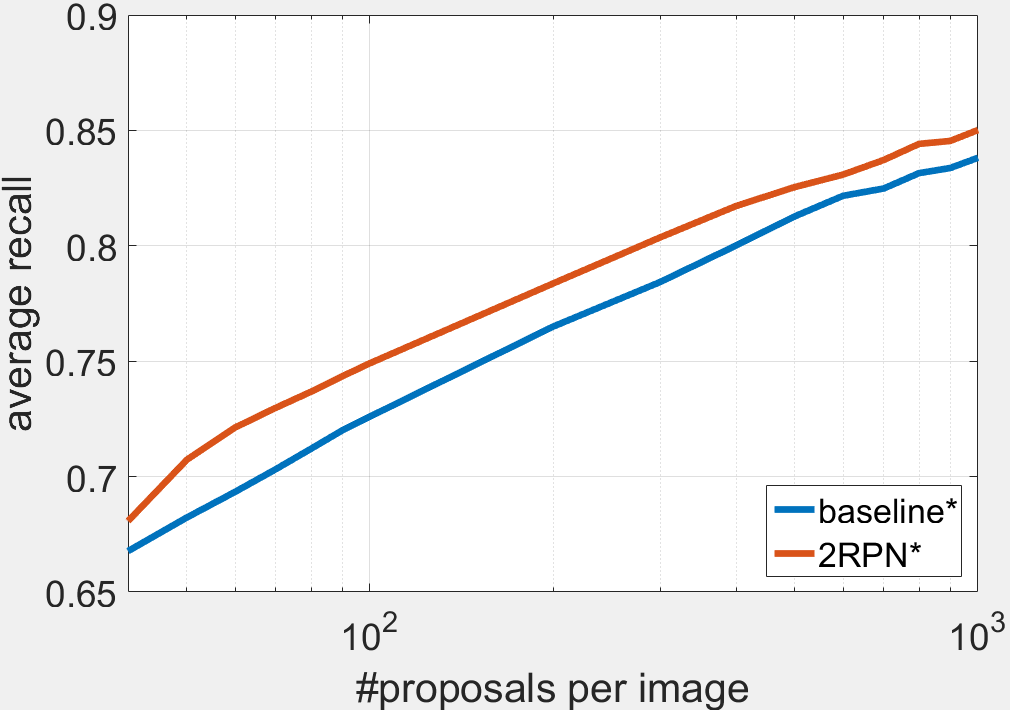}        
	\caption{Average recall (AR) curve with regard to \#proposals per image. AR is calculated on all pedestrians. Models are trained with \textbf{R+}.}
	\label{proposal quality}
\end{figure}

\noindent \textbf{Training Set.} Training set matters. When training with one subset, other pedestrians are treated as ignore regions, i.e. neither positive nor negative samples will be sampled in these areas. Comparing Table \ref{ablation study:0.3} with Table \ref{ablation study:0.65}, there exists a huge performance gap on \textbf{HO}. It is because the models in Table \ref{ablation study:0.65} are optimized purely on \textbf{R} and no pedestrians with heavy occlusion are used during training. It leads to limited generalization ability to Heavy set. But the situation is different when using \textbf{R+} for training, where the extra heavily occluded pedestrians are also used to optimize the parameters. Given the important role of training set, we mark it clearly for subsequent experiments and compare different methods under the same training settings.

\noindent \textbf{Parallel RPN.}  In HBAN, there are two RPN branches trained with different regions of a ground truth and produce proposals with different characteristics, which are expected to be complementary. In order to validate it, we train a network with parallel RPN and disable parallel RCNN or AlignLoss. The performance increases on \textbf{R}/\textbf{HO} for HBAN trained with \textbf{R} and \textbf{R+} are 0.60\%/3.63\% and 0.79\%/2.44\%, respectively. Since we mainly focus on the proposal quality here, we report average recall (AR) as a function of the number of proposals per image, in Figure \ref{proposal quality}. The curves are plotted for models trained with \textbf{R+} and AR is calculated on all pedestrians in validation set. AR of parallel RPN is consistently higher than baseline, with an average difference of 1.5 pp, which shows that the s-head RPN is able to produce some complementary proposals to body branch and therefore increase the performance.

\noindent \textbf{Parallel RCNN and AlignLoss.} First, as in previous experiment, we disable parallel RPN or AlignLoss and only adopt parallel RCNN.  The performance increases on \textbf{R}/\textbf{HO} for HBAN trained with \textbf{R} and \textbf{R+} are 0.47\%/3.17\% and 0.87\%/1.39\%, respectively, which come from multi-task-loss training detailed discussed in Section 3.3. When AlignLoss is further added in RCNN stage, the increases on \textbf{HO} for models trained with \textbf{R} and \textbf{R+} are 1.62\% and 1.98\%, which is mainly due to more accurately localized detections. From the relatively large increase on \textbf{HO}, it could be seen that localization accuracy is important under crowded scenes, which could make it easier to discriminate individuals in a crowd. 

\begin{figure}[t]
	\centering
	\includegraphics[width=1.0\linewidth, height=0.8\linewidth]{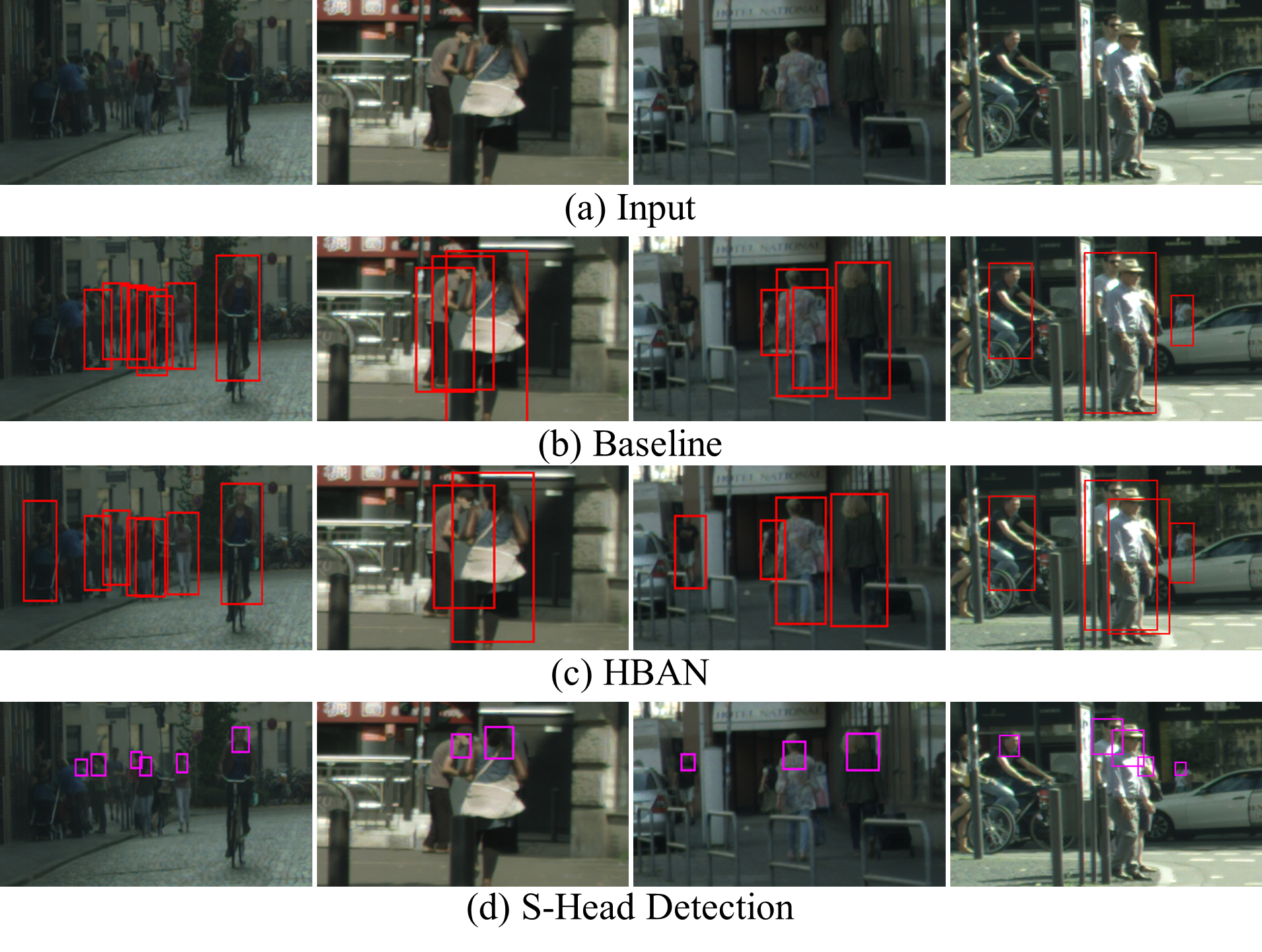}
	\vspace{-5mm}
	\caption{Visual comparisons. Compared with baseline, HBAN outputs more compact boxes and is able to produce detections when the body part is heavily occluded. The last row is the s-head detections output by HBAN. } 
	\label{fig:visual}
\end{figure}

\noindent \textbf{Full model.} The complete model reaches top performance for each subset, especially on \textbf{HO}, with about 5\% increase over baseline. In addition, some visual comparisons of the baseline and HBAN are shown in Figure \ref{fig:visual}. HBAN produces more correctly localized detections in a crowd than baseline. The s-head detection results are also shown in the figure and the network fails to produce some small s-heads. It is because the resolution of small heads is insufficient after pooling for accurate classification, which could be improved by using image up-scaling.

\noindent \textbf{Computation Overhead.} The s-head branch is in parallel with body. Therefore some calculations can be carried out at the same time. Single body branch takes 4.2GB memory and 0.51s to test a 2048*1024 image with 300 proposals. The complete model takes 5.6GB and 0.73s in comparison.

\begin{table}
	\begin{center}
		\renewcommand\arraystretch{1.1}
		\setlength{\tabcolsep}{2.5mm}{
		\begin{tabular}{@{}lcccc@{}}
			\hline
			Method & Data & Scale & \textbf{R} & \textbf{HO}   \\
			\hline
			\hline
			Adapted FRCNN\cite{citypersons} & \textbf{R} & $\times1$ & 15.4  & -  \\
			\hline
			Part+Grid \cite{improve-handling} & All & $\times1$ & 16.8 & 48.5  \\
			Att-part \cite{attention} & \textbf{R} & $\times1$ & 16.0 & 56.7 \\
			TLL(MRF) \cite{small} & - & $\times1$ & 14.4 & 52.0   \\
			OR-CNN \cite{ORCNN} & \textbf{R} & $\times1$ & 12.8 & 55.7  \\
			HBAN (ours) & \textbf{R} & $\times1$ & \textbf{12.5} & \textbf{48.1}  \\
			\hline
			RepLoss \cite{repulsion} & \textbf{R} & $\times1$ & 13.2 & 56.9  \\
			Adaptive NMS \cite{adaptive-NMS} &\textbf{R} & $\times1$ & \textbf{11.9} & 55.2 \\
			HBAN (ours) & \textbf{R} & $\times1$ & 12.5 & \textbf{48.1}  \\
			\hline
			RepLoss \cite{repulsion} & \textbf{R} & $\times1.3$ & 11.6 & 55.3  \\
			OR-CNN \cite{ORCNN} & \textbf{R} & $\times1.3$ & 11.0 & 51.3  \\
			HBAN (ours) & \textbf{R} & $\times1.3$ & \textbf{10.7} & \textbf{46.9}  \\
			\hline
		\end{tabular}
		}
	\end{center}
	\vspace{-4mm}
\caption{Comparison of HBAN with other methods on CityPersons validation set. `Data' represents the training set and `Scale' represents the scale of images used in training and testing. The best performances for each case are in bold.}
\label{compare citypersons: R}
\end{table}

\vspace{4mm}

\begin{table}
	\begin{center}
		\renewcommand\arraystretch{1.1}
		\setlength{\tabcolsep}{4.0mm}{
		\begin{tabular}{@{}lcccc@{}}
			\hline
			Method & Data & Scale & \textbf{R} & \textbf{HO}   \\
			\hline
			\hline
			Bi-Box \cite{bi-box} & \textbf{R+} & $\times1.3$ & 11.2 & 44.2 \\ 
			MGAN \cite{mask-guided} & \textbf{R+} & $\times1.3$ & \textbf{10.5} & \textbf{39.4} \\
			HBAN(ours) & \textbf{R+} & $\times1.3$ & 10.9 & 40.8 \\
			\hline
		\end{tabular}
		}
	\end{center}
	\vspace{-4mm}
	\caption{Comparison of HBAN with other methods on CityPersons validation set. Models trained with \textbf{R+} set are demonstrated.}
	\label{compare citypersons: R+}
\end{table}
			
\begin{table}
	\begin{center}
		\renewcommand\arraystretch{1.1}	
		\setlength{\tabcolsep}{3.0mm}{		
		\begin{tabular}{@{}lcccc@{}}
			\hline
			Method & Data & Scale & \textbf{R} & \textbf{HO}   \\
			\hline
			\hline
			ALF Net \cite{asymtotic-fitting} & \textbf{R} \& Aug & $\times1$ & 12.0 & 51.9   \\
			CSP \cite{CSP} & \textbf{R} \& Aug & $\times1$ & 11.0 &  49.3   \\
			HBAN (ours) & \textbf{R} \& Aug & $\times1$ & \textbf{10.9} & \textbf{47.0} \\
			\hline
		\end{tabular}
	}
	\end{center}
	\vspace{-4mm}
	\caption{Comparison of HBAN with other methods on CityPersons validation set. Models trained with additional data augmentation methods are demonstrated. Here `Scale' is only for testing phase.}
	\label{compare citypersons: Aug}
\end{table}

\begin{table}
	\begin{center}
		\renewcommand\arraystretch{1.1}
		\setlength{\tabcolsep}{5.0mm} {
			\begin{tabular}{@{}lcc@{}}
				\hline
				Method & \textbf{R} & \textbf{HO} \\
				\hline
				\hline
				Adapted FRCNN \cite{citypersons} & 12.97 & 50.47 \\
				RepLoss \cite{repulsion} & 11.48 & 52.59 \\
				Cascade MS-CNN \cite{cascade-ped} & 11.62 & 47.14 \\
				OR-CNN \cite{ORCNN} & 11.32 & 51.43 \\	
				MGAN \cite{mask-guided}	& \textbf{9.29} & 40.97 \\	
				\hline 
				HBAN (ours) & 11.26 & \textbf{39.54}\\
				\hline
			\end{tabular}
		}
	\end{center}
	\vspace{-3mm}
	\caption{Comparison of HBAN with other published state-of-the-art methods on CityPersons test set. }
	\label{compare citypersons test}
\end{table}

\subsection{Comparisons with State-of-the-art Methods}
We compare HBAN with other recent state-of-the-art pedestrian detectors on CityPersons dataset.  We report the performances of HBAN and other methods on the validation set in Table \ref{compare citypersons: R}, \ref{compare citypersons: R+} and \ref{compare citypersons: Aug}. The comparisons on CityPersons test set are reported in Table \ref{compare citypersons test}.

First, for part-based methods, we compare with OR-CNN \cite{ORCNN}, TLL \cite{small}, Att-part \cite{attention} and Part+Grid \cite{improve-handling}. All methods except Part+Grid use \textbf{R} set for training. Att-part also uses human key-point detection results to improve the performance. According to Table \ref{compare citypersons: R}, HBAN achieves the best performance on \textbf{R} set and outperforms Part+Grid on \textbf{HO} by using far less annotations. HBAN is especially effective on \textbf{HO} across all part-based methods.

RepLoss \cite{repulsion} and Adaptive NMS \cite{adaptive-NMS} are models designed to better discriminate individuals in a crowd. HBAN outperforms both methods on \textbf{HO} with a large margin. Image up-scaling is proved to be useful in OR-CNN and RepLoss and improves the performance significantly. HBAN also achieves the best performance when training with 1.3x up-scaled images.

For training with \textbf{R+} set, we compare with Bi-Box \cite{bi-box} and MGAN \cite{mask-guided}. Similar to HBAN, Bi-Box also adopts a parallel branches detection framework. It detects the visible boxes using additional supervision on the other branch while HBAN detects the semantic head with inferred supervision. MGAN predicts the visible mask of each pedestrian and produces attention weights.  Both Bi-Box and MGAN use the visible box annotation as supervision to the network, while HBAN only use it to filter out poor positive samples. HBAN produces comparable results with MGAN and outperforms Bi-Box on both \textbf{R} and \textbf{HO} sets.

Moreover, ALF Net \cite{asymtotic-fitting} and CSP \cite{CSP} adopt other data augmentation methods including color distortion, image cropping and resizing. Similarly, we also use color distortion and random scaling in training HBAN and the results are displayed in Table \ref{compare citypersons: Aug}. Compared with the performances reported in Table \ref{compare citypersons: R}, the increase is 1.6\% and 1.1\% on \textbf{R} and \textbf{HO}, respectively and HBAN outperforms both methods with a relatively large margin.

On the test set, we submit our results trained with $\times1.3$ up-scaled images from \textbf{R+} set for on-line evaluation and compare with other published methods in Table \ref{compare citypersons test}. HBAN achieves the best performance on \textbf{HO}. 

From the above analysis, Table \ref{compare citypersons: R} demonstrates the effectiveness of HBAN compared with other part-based methods, especially for heavily occluded pedestrians. Table \ref{compare citypersons: R+} and \ref{compare citypersons: Aug} further shows the potential of HBAN when using more training information and data augmentation. HBAN reaches state-of-the-art performance on CityPersons validation and test set, demonstrating its superiority under occlusion.

\section{Conclusion and Discussion}
\vspace{-1mm}
Head is an important cue for pedestrian detection in a crowd because of its stable appearance, visibility and relative position. Inspired by it, we propose semantic head detection in parallel with body branch to address intra-class occlusion. The semantic head detector is trained using inferred annotations whose positions are fixed relative to body boxes. 
Further, AlignLoss is designed to explicitly restrict the relation between s-head and body. Detailed experiments have demonstrated the effectiveness of the proposed HBAN under various occlusion levels. 

In this paper, we show that using a special part of full body as supervision could improve the detection performance of full body. While it is true for pedestrian, whose structure is relatively stable and easy to predict, we are looking forward to generalizing it into common object detection. Further more, as HBAN is able to produce head detections, it is also interesting to explore the usage of semantic head in post processing stages, like fusing s-head detections into NMS.

\section*{Acknowledgments}
This work was supported by the National Key R\&D Plan (No. 2016YFB0100901), the National Natural Science Foundation of China (No. 61171113 and No. 61773231).








\section*{Reference}
\bibliography{mybibfile}
\bibliographystyle{elsarticle-num}

\end{document}